\documentclass[letterpaper, 10 pt, conference]{ieeeconf}

\IEEEoverridecommandlockouts
\usepackage[noadjust]{cite}

\makeatletter
\let\NAT@parse\undefined
\makeatother

\usepackage{kotex}

\usepackage[dvipsnames]{xcolor}
\usepackage[colorlinks=true, linkcolor=blue, citecolor=blue, urlcolor=blue, pagebackref=true]{hyperref}
\definecolor[named]{ACMDarkBlue}{cmyk}{1,0.58,0,0.21}
\definecolor{darkgreen}{RGB}{25,200,25}
\hypersetup{colorlinks,
  linkcolor=ACMDarkBlue,  % ACMPurple,
  citecolor=ACMDarkBlue,  % ACMPurple,
  urlcolor=ACMDarkBlue,   % black,
  filecolor=ACMDarkBlue
}

\usepackage{caption}
\DeclareCaptionFormat{tablewithgap}{#1#2\vspace{3pt}#3}
\usepackage{tabularx,booktabs}
\newcolumntype{Y}{>{\centering\arraybackslash}X}
\usepackage{colortbl}
\usepackage{multirow}
\usepackage{soul}
\usepackage{wrapfig}

\usepackage{amsmath}
\usepackage{amssymb}
\usepackage{tikz}
\usetikzlibrary{shapes.geometric, arrows.meta, positioning, fit, backgrounds}

\let\labelindent\relax
\usepackage{enumitem}

\usepackage{tikz}

\newcommand\preprinttext{%
    \scriptsize This is a preprint - Published in IEEE/RSJ International Conference on Intelligent Robots and Systems (IROS) 2026
}
\newcommand\copyrighttext{%
    \scriptsize \textcopyright 2026 IEEE. Personal use of this material is permitted. Permission from IEEE must be obtained for all other uses, in any current or future media, including reprinting/republishing this material for advertising or promotional purposes, creating new collective works, for resale or redistribution to servers or lists, or reuse of any copyrighted component of this work in other works.
}
\newcommand\noticeblock{%
    \begin{tikzpicture}[remember picture,overlay]
    \node[anchor=north,yshift=-30pt] at (current page.north) {\preprinttext};
    \node[anchor=south,yshift=20pt] at (current page.south)
      {\parbox{\dimexpr\textwidth-\fboxsep-\fboxrule\relax}{\copyrighttext}};
    \end{tikzpicture}%
}

\newcommand{\model}{\textit{3D HAMSTER}}
\newcommand{\benchmark}{\textit{DroidSpatial-Bench}}

\title{\LARGE \bf
3D HAMSTER: Bridging Planning and Control in Hierarchical Vision Language Action Models through 3D Trajectory Guidance
}

\author{Dongyoon Hwang$^{1*}$, Byungkun Lee$^{1*}$, Dongjin Kim$^{1*}$, Hyojin Jang$^{1}$, Hoiyeong Jin$^{1}$, Jueun Mun$^{2}$, \\ Minho Park$^{1}$, Hojoon Lee$^{3}$, Hyunseung Kim$^{1, 4}$, and Jaegul Choo$^{1\dag}$% 
\thanks{$^{*}$Equal contribution, $^{\dag}$Corresponding author.}% 
\thanks{This research was supported by a grant from KRAFTON AI and the “Advanced GPU Utilization Support Program” funded by the Government of the Republic of Korea (Ministry of Science and ICT).}%
\thanks{$^{1}$Dongyoon Hwang, Byungkun Lee, Dongjin Kim, Hyojin Jang, Hoiyeong Jin, Minho Park, Hyunseung Kim, and Jaegul Choo are with the Kim Jaechul Graduate School of AI, KAIST, Seoul, Republic of Korea (e-mail: godnpeter@kaist.ac.kr).}%
\thanks{$^{2}$Jueun Mun is with the Graduate School of Artificial Intelligence, POSTECH, Pohang, Republic of Korea.}%
\thanks{$^{3}$Hojoon Lee is with Holiday Robotics, Seoul, Republic of Korea.}%
\thanks{$^{4}$Hyunseung Kim is with KRAFTON AI, Seoul, Republic of Korea.}%
\thanks{Links: \href{https://github.com/DAVIAN-Robotics/3D_HAMSTER}{GitHub Code} $\vert$ \href{https://huggingface.co/DAVIAN-Robotics/3D_HAMSTER}{HF Models} $\vert$ \href{https://davian-robotics.github.io/3D_HAMSTER/}{Project Page}}%
}

\begin{document}
\bstctlcite{BSTcontrol}
\maketitle
\thispagestyle{empty}
\pagestyle{empty}
\noticeblock

%%%%%%%%%%%%%%%%%%%%%%%%%%%%%%%%%%%%%%%%%%%%%%%%%%%%%%%%%%%%%%%%%%%%%%%%%%%%%%%%

\begin{abstract}
Hierarchical Vision-Language-Action (VLA) models decouple high-level planning from low-level control to improve generalization in robot manipulation. Recent work in this paradigm uses 2D end-effector trajectories predicted by a Vision-Language Model (VLM) as explicit guidance for a downstream policy. However, state-of-the-art low-level policies operate in 3D metric space on point clouds, and feeding them 2D guidance that lacks depth forces each waypoint to be assigned the depth of whatever scene surface lies beneath it, producing geometrically distorted trajectories. We propose \model{}, a hierarchical framework that closes this gap by having the planner directly output metrically reliable 3D trajectories. We augment a VLM with a dedicated depth encoder and a dense depth reconstruction objective to predict 3D waypoint sequences, which are directly integrated into a pointcloud-based low-level policy. Across 3D trajectory prediction, simulation, and real-world manipulation, \model{} consistently outperforms proprietary VLMs and 2D-guided baselines, with the largest gains under appearance-altering shifts and unseen language, spatial, and visual conditions. The project page is available at \url{https://davian-robotics.github.io/3D_HAMSTER/}. 
\end{abstract}

% %%%%%%%%%%%%%%%%%%%%%%%%%%%%%%%%%%%%%%%%%%%%%%%%%%%%%%%%%%%%%%%%%%%%%%%%%%%%%%%%

\section{INTRODUCTION}

A long-standing challenge in robot manipulation is bridging the gap between high-level semantic reasoning and low-level motor control. The strong semantic understanding capabilities of Vision-Language Models (VLMs)~
\cite{beyer2024paligemma, bai2025qwen3vl, alayrac2022flamingo} have inspired the development of end-to-end Vision-Language-Action (VLA) architectures that directly map visual observations and language instructions to continuous actions~\cite{zitkovich2023rt2, kim2024openvla, black2025pi05, bjorck2025gr00t}.
However, end-to-end VLA architectures, also referred to as \textit{monolithic} models~\cite{kim2024openvla,black2025pi05,bjorck2025gr00t}, have shown limited performance due to the scarcity of robot demonstration data, which remains costly to collect at scale on physical hardware.
Fine-tuning on this limited data erodes the broad generalizability of the underlying VLM, leaving these models vulnerable to out-of-distribution (OOD) visual shifts across novel objects, viewpoints, and environments~\cite{ fei2025liberoplus}.

To better leverage the generalization capability of VLMs, hierarchical VLA frameworks~\cite{li2025hamster, ma2026generalvla, huang2025thinkact} have been proposed as a compelling alternative that explicitly decouples semantic reasoning from low-level motor control.
Specifically, they utilize a VLM as a high-level planner that produces 2D keypoints on a camera image (System~2), while a low-level controller receives this guidance to generate motor commands (System~1).
This decoupling provides a critical scaling advantage. Because the planner predicts visual targets rather than robot-specific actions, it can be trained on abundant non-robot data encompassing spatial reasoning, visual grounding, 2D bounding boxes, and general VQA, preserving the broad generalizability of the underlying VLM.

\begin{figure}[t]
    \centering
    \includegraphics[width=0.9\columnwidth]{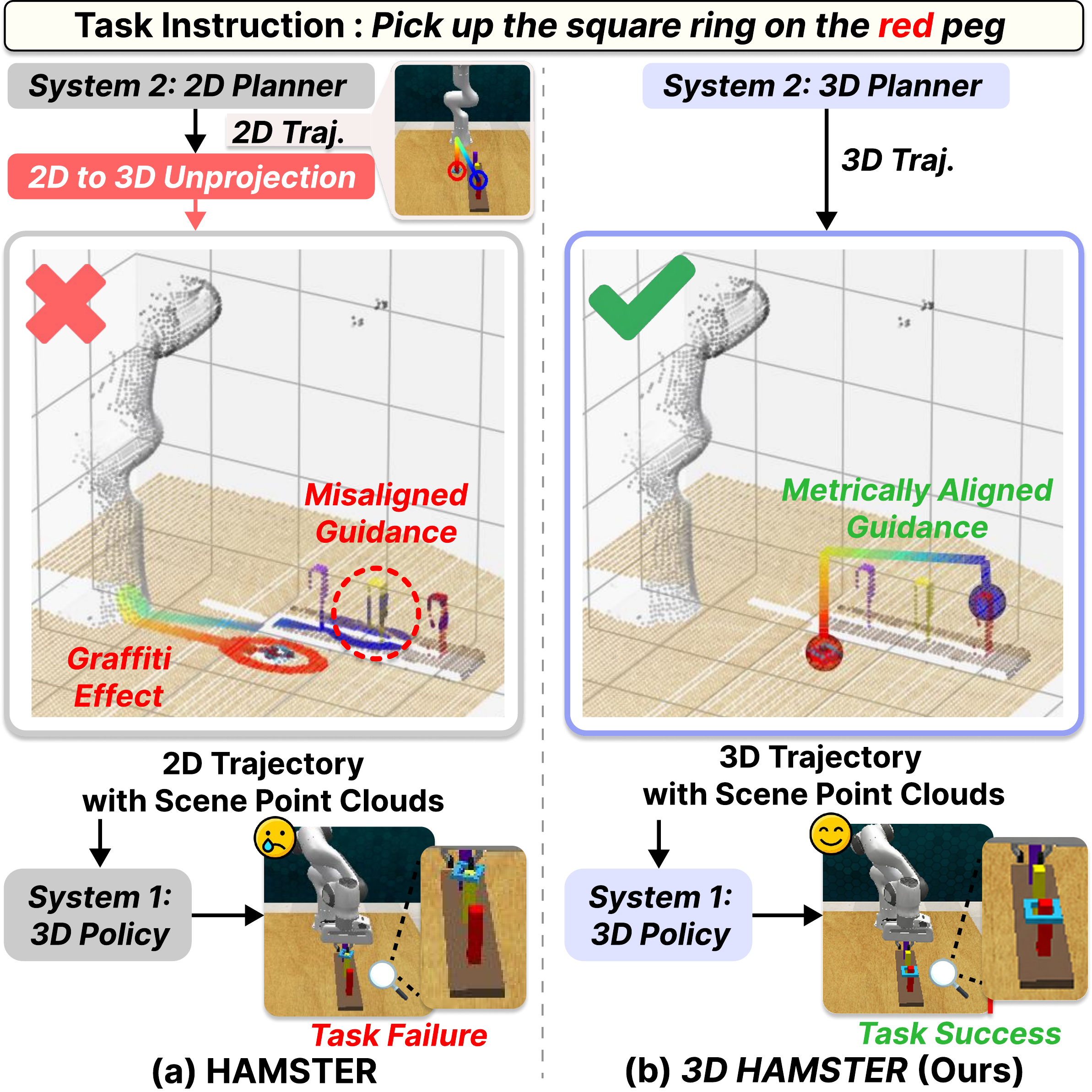}
    \caption{Comparison of 2D and 3D guidance in hierarchical VLAs. (a) 2D planners create representational misalignment: unprojecting 2D plans yields flawed 3D guidance, leading to brittle execution. (b) Our 3D-aware planner generates metrically reliable 3D guidance, establishing a shared metric space for robust manipulation.}
    \label{fig:1}
    \vspace{-0.7cm}
\end{figure}

While this 2D formulation is a natural fit for VLM-based planners, recent research on low-level policies has increasingly favored 3D-native architectures that operate on point clouds, consistently outperforming 2D alternatives in spatial precision and robustness to viewpoint changes~\cite{ze20243ddp, gervet2023act3d, ke20243ddiffuseractor, gkanatsios20253dfa}. This creates a fundamental representational misalignment in current hierarchical VLAs: the planner would reason in 2D pixel coordinates, while the controller operates in 3D space. 

For instance, HAMSTER~\cite{li2025hamster}, pairs a 2D VLM planner with a pointcloud-based 3D controller (Fig.~\ref{fig:1}a). To bridge the gap between 2D plans and 3D execution, the planner's 2D waypoints must be lifted into 3D by sampling depth from the scene surface at each pixel location. However, since the planner itself provides no depth information, the result is determined entirely by whatever geometry happens to lie beneath each waypoint.
This produces a \textit{graffiti effect} (Fig.~\ref{fig:1}a): the trajectory clings to the scene surface rather than passing freely through 3D space, making it difficult for the controller to distinguish the intended path from the geometry itself.
If, instead, the planner directly outputs 3D coordinates, no such projection is needed: the trajectory exists in the same metric space as the controller from the start, faithfully reflecting the planner's intent (Fig.~\ref{fig:1}b).

Fortunately, recent VLMs have acquired substantial 3D spatial knowledge through large-scale geometry-rich pretraining, making 3D-aware planning increasingly feasible~\cite{bai2025qwen3vl, hu2025g2vlm}.
Building on this foundation, we propose \model{}, a hierarchical VLA framework that trains a VLM backbone to predict metrically reliable 3D end-effector trajectories in $(u,v,d)$ form, where $(u, v)$ are image-plane coordinates and $d$ is metric depth, producing guidance directly aligned with the operating space of 3D low-level policies. Since RGB features alone cannot reliably recover metric depth from a single view, we augment the VLM with a dedicated depth encoder and introduce a dense depth reconstruction objective that regularizes the model's internal representations to preserve faithful scene geometry, complementing sparse trajectory supervision with a scene-level 3D prior. To maintain broad generalization, we train the planner on a mixture of 3D capability data and 2D preservation data. The predicted trajectories are then transformed into world coordinates and executed by a pointcloud-based low-level policy, ensuring a unified metric interface from planning through execution.

We evaluate \model{} across three settings. (1) On \benchmark{}, a benchmark we construct from held-out DROID pick-and-place episodes~\cite{khazatsky2024droid}, our depth-encoder-augmented VLM produces more accurate 3D trajectories than strong baselines including Gemini-3.0-Pro and RoboBrain~2.5. (2) On 11 tasks from the Colosseum simulation benchmark~\cite{pumacay2024colosseum}, which stress-tests policies under 14 perturbation axes, 3D trajectory guidance consistently outperforms 2D alternatives, with the largest gains under appearance-altering shifts such as lighting and texture changes. (3) On a real Franka Panda arm across three tasks of increasing spatial precision (button pressing, pouring, and pick-and-place), 3D guidance improves performance across four generalization axes: unseen language, spatial references, visual conditions, and their combination. Our contributions are as follows:
\begin{itemize}[leftmargin=1.0em]
    \item A framework that closes the 2D-3D representational gap in hierarchical VLAs by having the planner output 3D trajectories directly consumable by pointcloud-based policies.
    \item A training recipe combining a depth encoder, dense reconstruction loss, and a curated data mixture that enables a pretrained VLM to generate metric 3D trajectories.
    \item Experimental validation across three complementary settings showing that aligning planning and execution in 3D metric space yields robust manipulation under distribution shifts where 2D guidance fails.
\end{itemize}

\section{RELATED WORK}

\subsection{Hierarchical Vision-Language-Action (VLA) Models}

Existing hierarchical VLA frameworks decouple high-level planning from low-level control through VLM-friendly intermediate representations such as language~\cite{wen2025dexvla}, 2D keypoints~\cite{yuan2024robopoint}, and 2D end-effector
trajectories~\cite{li2025hamster, huang2025thinkact, ma2026generalvla}. However, since fine-grained manipulation fundamentally operates in 3D space, these 2D representations introduce ambiguity under 3D-sensitive shifts such as occlusion or viewpoint changes. This limitation has motivated a shift toward 3D-aware planning. Concurrent to our work, RoboTracer~\cite{zhou2025robotracer} and RoboBrain~2.5~\cite{tan2026robobrain25} take a step in this direction by
producing depth-aware keypoint sequences in $(u,v,d)$ form. RoboTracer further incorporates depth input and predicts metric scale via a regression-supervised decoder, but remains a standalone motion planner that does not integrate its predicted trajectories with a downstream controller for closed-loop manipulation. In contrast, \model{} enforces scene-level geometric understanding through a dense depth reconstruction loss and investigates how to effectively couple the predicted 3D trajectories with a pointcloud-based low-level policy, closing the full loop from 3D-aware planning to 3D-native execution.

\begin{figure*}[t]
    \centering
    \includegraphics[width=\textwidth]{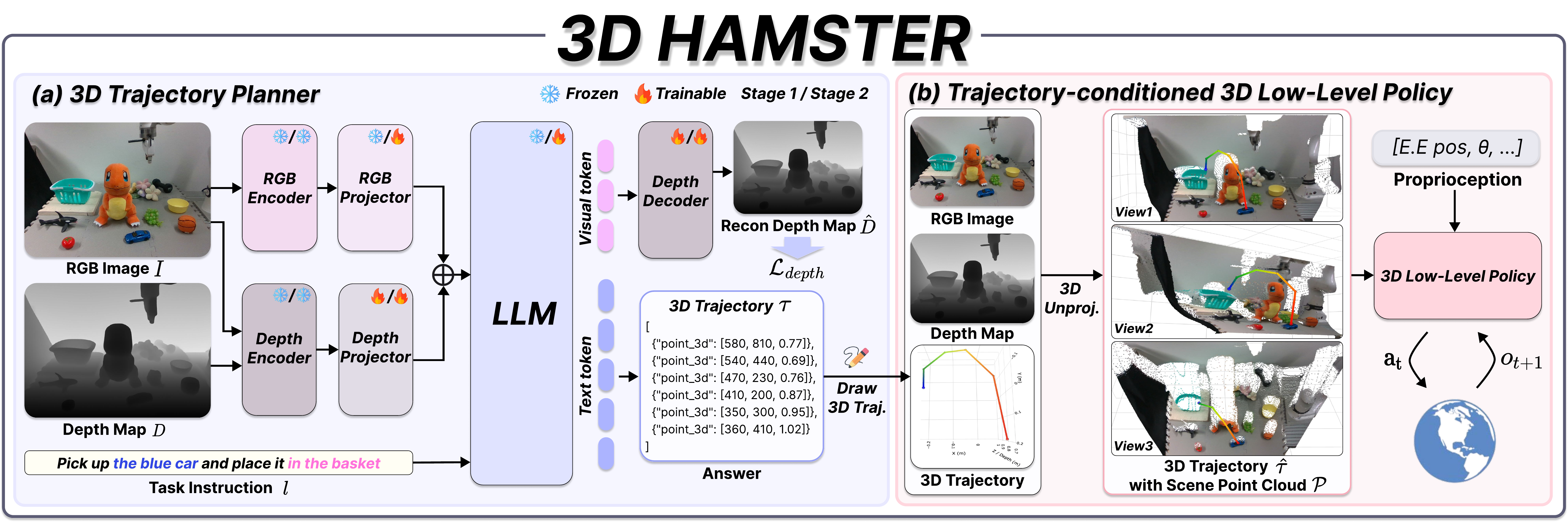}
    \caption{Overview of \model{}. The framework decouples semantic planning and motor execution with two-stage training strategy: Stage 1 aligns depth features with the VLM space using a dense reconstruction loss ($\mathcal{L}_{depth}$) while preserving VLM capabilities; Stage 2 fine-tunes for trajectory prediction. The 3D trajectory planner fuses RGB and depth to generate metrically reliable 3D trajectories, which the trajectory-conditioned 3D low-level policy executes as robust closed-loop actions ($\mathbf{a_t}$) from the scene point cloud.}
    \label{fig:2}
    \vspace{-0.7cm}
\end{figure*}

\subsection{Vision-Language Models (VLMs) for Spatial Reasoning}
VLMs are increasingly used for spatial prediction in embodied robotics, with outputs evolving from point estimates~\cite{nasiriany2024pivot,sundaresan2023kite, yuan2024robopoint, fangandliu2024moka,
zhou2026roborefer} to trajectory-level guidance~\cite{gu2023rttrajectory, zhou2025robotracer,tan2026robobrain25}. Beyond 2D spatial reasoning, recent VLMs have also acquired rough 3D spatial knowledge through large-scale pretraining on
geometry-rich data, as evidenced by capabilities such as 3D bounding box prediction in Qwen3-VL~\cite{bai2025qwen3vl} and joint 3D scene reconstruction and spatial question answering in G$^2$VLM~\cite{hu2025g2vlm}. Other works go further by injecting
explicit geometric priors into VLMs through 3D grounding data~\cite{wang2025n3dvlm}, 3D geometry encoders~\cite{zheng2025vgllm}, or 3D vision-language-action instruction tuning~\cite{huang2024leo}. These results collectively demonstrate that modern VLMs possess a meaningful foundation of 3D spatial understanding. However, prior robotic planning methods have not fully leveraged this
pretrained 3D knowledge, largely confining VLM planners to 2D outputs. Our work shows that this latent 3D capability can be effectively channeled into metrically reliable trajectory prediction for robotic manipulation through targeted fine-tuning and architectural augmentation.

\subsection{3D Low-Level Manipulation Policies}
Earlier manipulation policies such as Diffusion Policy~\cite{chi2025diffusion} operate on 2D image observations and
struggle with depth ambiguities and perspective distortions in contact-rich tasks~\cite{diffusionpolicyseurvey}. The field has shifted toward 3D-native architectures that operate directly on metric representations such as point clouds. DP3~\cite{ze20243ddp} adopts pointcloud-based state representations within diffusion-based frameworks. Act3D~\cite{gervet2023act3d} and RVT-2~\cite{goyal2024rvt2} leverage 3D feature fields and multi-view rendering for precise spatial prediction. 3D Diffuser Actor~\cite{ke20243ddiffuseractor} unifies diffusion policies with 3D scene representations through a relative position denoising transformer. 3D FlowMatch Actor (3DFA)~\cite{gkanatsios20253dfa} extends 3DDA by replacing diffusion with rectified flow matching for faster and more efficient end-effector trajectory prediction. The consistent gains of these 3D controllers motivate our choice of a pointcloud-based low-level policy, and more broadly, the need for a high-level planner that produces guidance in the same 3D space.

\section{METHOD}
\label{sec:method}

\subsection{\model}
\label{sec:method:formulation}

\paragraph{Problem Formulation}

Given calibrated RGB-D cameras and a natural language instruction, our goal is to produce motor actions that accomplish the described manipulation task. At each timestep, the robot observes an RGB image $I \in \mathbb{R}^{H \times W \times 3}$, a depth map $D \in \mathbb{R}^{H \times W}$, and a task instruction $l$. We decompose this into a high-level planner that predicts an end-effector trajectory $\tau = \{(u_t, v_t, d_t)\}_{t=1}^{T}$ in pixel coordinates $(u_t, v_t)$ with metric depth $d_t$, where $T$ is the number of waypoints and $t$ indexes each waypoint along the trajectory, and a low-level policy $\pi_{\text{low}}(\mathcal{P}, \tau) = \{\mathbf{a_t}\}_{t=1}^{T_a}$ that produces a chunk of $T_a$ actions from the current point cloud observation $\mathcal{P}$ conditioned on $\tau$, executing closed-loop control at each re-planning step.

\paragraph{Overall Framework}
As illustrated in Fig.~\ref{fig:2}, \model{} consists of a 3D trajectory planner and a trajectory-conditioned 3D low-level policy. The 3D trajectory planner (Fig.~\ref{fig:2}a) takes a single RGB image $I$, a depth map $D$, and a task instruction $l$ as input, and autoregressively generates a metrically reliable end-effector trajectory $\tau$. To produce geometrically faithful predictions, we augment the VLM with a dedicated depth encoder that produces depth features alongside the RGB features from the vision encoder. These two feature sets are then fused into unified visual tokens and passed to the LLM backbone, which is further regularized by a dense depth reconstruction loss to preserve metrically accurate scene geometry. The trajectory-conditioned 3D low-level policy (Fig.~\ref{fig:2}b) then unprojects $\tau$ into world coordinates using known camera intrinsics and extrinsics, appends the resulting 3D waypoints to the scene point cloud constructed from RGB-D observations, and predicts a chunk of $T_a$ actions $\{\mathbf{a_t}\}_{t=1}^{T_a}$ through rectified flow matching~\cite{liu2022flow} over this unified 3D representation.

\subsection{3D Trajectory Planner}
\label{sec:method:sys2}

We build on Qwen3-VL~\cite{bai2025qwen3vl}, which possesses broad 3D spatial knowledge from large-scale pretraining, including the ability to predict 3D bounding boxes from a single image. However, this knowledge was acquired through object-level localization tasks and does not directly transfer to generating coherent sequences of 3D waypoints for end-effector trajectory prediction. Without targeted adaptation, the base model achieves near-zero accuracy on this task (Table~\ref{tab:traj_prediction}, row 1 of \textit{ours}). We bridge this gap through three complementary components: (i) a curated training mixture, (ii) a dedicated depth encoder, and (iii) a dense depth reconstruction loss, whose individual contributions we isolate in Sec.~\ref{sec:exp:rq1}.

\paragraph{Training Data}
We train the planner on a mixture of eight data sources grouped into two categories (Table~\ref{tab:system2_training_data}), designed to channel the base VLM's 3D knowledge into metric trajectory prediction without eroding its existing vision-language understanding. \textit{(i) 3D capability data} (RGB + depth) consists of end-effector trajectories from real and simulated robot demonstrations and spatial reasoning data from RefSpatial~\cite{zhou2026roborefer}. Each trajectory sample pairs an RGB image and depth map with a $(u, v, d)$ waypoint sequence and gripper actions. We include three supervision variants: 2D-only, 3D-only, and a chain-of-thought variant that first predicts the 2D trajectory and then lifts it to 3D, encouraging the model to learn consistent pixel-to-depth mappings. RefSpatial, originally a 2D-only dataset, is augmented with generated depth maps and extended to include depth-aware spatial reasoning and vacant-space localization, providing additional supervision for the depth encoder. Ground-truth depth is used for RLBench~\cite{james2020rlbench} and DROID~\cite{khazatsky2024droid}; for InternData-M1~\cite{internrobotics2025interndata-m1} and RefSpatial, we generate metric depth using MoGe-2~\cite{wang2025moge2}. 
\textit{(ii) Preservation data} (RGB only) anchors the base model's original vision-language capabilities, preventing them from being overwritten during trajectory fine-tuning. This includes RoboPoint~\cite{yuan2024robopoint} for point-based referring, PixMo~\cite{deitke2025molmo} for indoor pointing, LVIS~\cite{gupta2019lvis} for bounding-box detection, and Honey-1M~\cite{zhang2025honeydata} for general VQA. These samples are processed entirely through the base VLM, bypassing the depth pathway.

\begin{table}[t]
  \centering
  \caption{3D Trajectory Planner Training Data Composition.}
  \label{tab:system2_training_data}
  \setlength{\tabcolsep}{4pt}
  \footnotesize
  \begin{tabular}{@{}llcll@{}}
    \toprule
    \textbf{Source} & \textbf{Env.} & \textbf{Modality} & \textbf{Tasks} & \textbf{Size} \\
    \midrule
    \multicolumn{5}{@{}l}{\textit{3D capability data}} \\[2pt]
    RLBench~\cite{james2020rlbench}          & Sim  & RGB-D & 2D / 3D / 2D$\to$3D  & 606K \\
    DROID~\cite{khazatsky2024droid}          & Real & RGB-D & 2D / 3D / 2D$\to$3D  & 123K \\
    InternData-M1~\cite{internrobotics2025interndata-m1}   & Sim  & RGB-D & 2D / 3D / 2D$\to$3D   & 1.5M \\
    RefSpatial~\cite{zhou2026roborefer}      & Mix & RGB-D & Spatial QA / Vacant loc.   & 2.2M \\
    \midrule
    \multicolumn{5}{@{}l}{\textit{Preservation data}} \\[2pt]
    RoboPoint~\cite{yuan2024robopoint}       & Sim & RGB   & 2D pointing           & 666K \\
    PixMo~\cite{deitke2025molmo}             & Real & RGB   & 2D pointing            & 171K \\
    LVIS~\cite{gupta2019lvis}                                     & Real & RGB   & 2D Bbox det.           & 138K \\
    Honey-1M~\cite{zhang2025honeydata}       & Web  & RGB   & General VQA                & 749K \\
    \bottomrule
  \end{tabular}
  \vspace{-0.5cm}
\end{table}

\paragraph{Depth Encoder}
While the training data provides supervision for metric depth prediction, RGB features alone cannot reliably recover metric depth from a single view. We therefore augment the VLM with a dedicated depth pathway (Fig.~\ref{fig:2}a). The RGB image $I$ is encoded into visual tokens by the pretrained visual encoder. 
In parallel, the depth map $D$ is processed by a separately initialized depth encoder to produce depth tokens that capture geometric cues complementary to the RGB features.
Both token sequences are projected into the LLM embedding space through their own dedicated projectors, fused element-wise, and then passed to the transformer backbone alongside the tokenized instruction $l$.
Given this fused visual, depth, and language context, the model autoregressively generates a trajectory $\tau = \{(u_t, v_t, d_t)\}_{t=1}^{T}$.

\paragraph{Depth Reconstruction Loss}
Simply providing depth tokens does not guarantee that the LLM's hidden states retain metrically faithful depth information. Since the model outputs the trajectory as a sequence of text tokens, relying solely on the standard autoregressive language modeling loss $\mathcal{L}_{\text{LM}}$ provides only a sparse gradient signal, which may cause the depth encoder's geometric features to degrade during fine-tuning. To prevent this, we route the depth tokens $z_D$ through a lightweight decoder $f_{\text{dec}}$ to reconstruct the full depth map $\hat{D} = f_{\text{dec}}(z_D)$, supervised by an $\ell_1$ loss:
\begin{equation}
  \mathcal{L}_{\text{depth}} = \|D - \hat{D}\|_1.
\end{equation}
The total training loss combines autoregressive trajectory prediction with this dense reconstruction signal:
\begin{equation}
  \mathcal{L} = \mathcal{L}_{\text{LM}} + \lambda\,\mathcal{L}_{\text{depth}},
\end{equation}
where $\lambda = 0.1$. While trajectory supervision is sparse and task-specific, the reconstruction loss provides a scene-level geometric prior that encourages metrically consistent depth understanding across varying objects and viewpoints.

\paragraph{Two-stage Training}
To integrate the depth pathway without disrupting the pretrained VLM, we train in two stages (Fig.~\ref{fig:2}a). In stage~1 (depth alignment stage), we freeze the RGB encoder, depth encoder, and LLM backbone, training only the depth projector and the depth decoder. This aligns the depth representation with the pretrained visual-language space while preserving
existing capabilities. In stage~2 (task fine-tuning stage), we freeze both the RGB and depth encoders to retain their learned representations and fine-tune the remaining parameters for trajectory prediction.

\begin{figure*}[t]
    \centering
    \includegraphics[width=\linewidth]{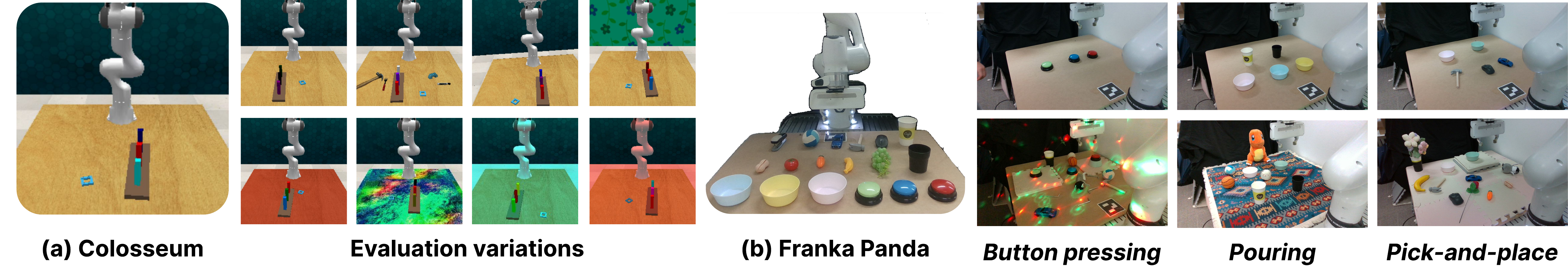}
    \caption{End-to-end manipulation evaluation setups. (a) Simulation environments from the Colosseum benchmark, showcasing various visual and physical perturbations. (b) Real-world setup using a Franka Panda arm equipped with an external RGB-D camera}
    \label{fig:4}
    \vspace{-0.5cm}
\end{figure*}

\subsection{Trajectory-conditioned 3D Low-level Policy}
\label{sec:method:sys1}

We adopt 3DFA~\cite{gkanatsios20253dfa}, a pointcloud-based policy that predicts actions via rectified flow matching, as our low-level backbone. The planner's trajectory~$\tau = \{(u_t, v_t, d_t)\}_{t=1}^{T}$ is first unprojected to a world coordinate trajectory $\hat{\tau} = \{(X_t, Y_t, Z_t)\}_{t=1}^{T}$ using known camera intrinsic matrix $K$ and extrinsic parameters $R, \mathbf{t_{cam}}$:
\begin{equation}
  \mathbf{p}_{\text{cam}} = d \cdot K^{-1} [u, v, 1]^\top, \quad
  \mathbf{p}_{\text{world}} = R\,\mathbf{p}_{\text{cam}} + \mathbf{t_{cam}}.
\end{equation}
The central challenge is how to condition this policy on the resulting 3D trajectory so that guidance and scene geometry operate in a shared representation.

\paragraph{Trajectory-Scene Fusion}
The world-frame trajectory~$\hat\tau$ is appended to the scene point cloud $\mathcal{P}$ (Fig.~\ref{fig:2}b), so that trajectory guidance and visual observations coexist in a unified 3D representation. Each waypoint is color-coded by its temporal position along the trajectory, enabling the policy to distinguish the progression of the planned motion. To further allow the policy to learn distinct strategies for following the trajectory versus understanding the scene, we add learnable modality embeddings to each point feature:

\begin{equation}
  \tilde{f}_j = f_j + \begin{cases} e_{\text{traj}} & \text{if }
  j \in \mathcal{I}_{\text{traj}}, \\ e_{\text{scene}} & \text{otherwise},
  \end{cases}
\end{equation}
where $f_j$ is the feature of point $j$, $\mathcal{I}_{\text{traj}}$ is the set of trajectory point indices, and $e_{\text{traj}}, e_{\text{scene}}$ are learned embeddings. Finally, since the policy subsamples the full point cloud for computational efficiency, we enforce that all trajectory points are preserved during subsampling by replacing the lowest-priority scene points with any dropped trajectory points. This guarantees that the guidance signal is never lost regardless of scene complexity.

\paragraph{Action Prediction}
Given the modality-tagged point cloud, the policy learns a velocity field $v_\theta$ under a rectified flow matching
formulation. A noised action sequence $\mathbf{a}^i = (1{-}i)\,\mathbf{a}^0 + i\,\epsilon$, where $\epsilon \sim \mathcal{N}(0, I)$ and $i \in [0,1]$, is mapped back to the clean action sequence $\mathbf{a}^0$. The training objective is:
\begin{equation}
  \mathcal{L}_{\text{policy}} = \|v_\theta(\mathbf{a}^i, \tilde{f}, i)
  - (\mathbf{a}^0 - \epsilon)\|_2^2 + \mathcal{L}_{\text{BCE}}(g_\theta, g^*),
\end{equation}
where the first term supervises the velocity field and the second supervises the gripper open/close action. At inference, the policy denoises from Gaussian noise to a clean action chunk of $T_a$ steps. Since trajectory guidance and scene share a common metric coordinate frame, the controller executes the plan directly without the implicit 2D-to-3D lifting required by 2D guidance approaches~\cite{li2025hamster}. 

% ============================================================
% EXPERIMENTS (Revised)
% ============================================================
\section{EXPERIMENTS}
\label{sec:experiments}

Our experiments address the following research questions:
\begin{itemize}[leftmargin=1.0em]
\item \textit{RQ1:} How important is depth for 3D end-effector trajectory prediction, and what adaptations enable a VLM to generate metric waypoints?
\item \textit{RQ2:} Does 3D trajectory guidance improve manipulation robustness under controlled visual and physical perturbations in RLBench?
\item \textit{RQ3:} Do the benefits of 3D guidance transfer to real-world manipulation across diverse generalization axes?
\end{itemize}

\subsection{Experimental Setup}

\paragraph{Evaluation} We evaluate in three settings: 3D trajectory prediction, simulation, and real-world manipulation.

For trajectory prediction (\textit{RQ1}), we propose \benchmark{}, constructed from 148 held-out DROID~\cite{khazatsky2024droid} pick-and-place episodes. Given an RGB-D image and a language instruction, the model predicts a 3D trajectory. We evaluate whether the predicted grasp and placement points each fall within $\delta \in \{5, 10\}$\,cm of ground truth, and report start-position, end-position, and combined (\emph{Both}) accuracies.

For simulated manipulation (\textit{RQ2}), we use the Colosseum benchmark~\cite{pumacay2024colosseum} (Fig.~\ref{fig:4}a), which extends RLBench~\cite{james2020rlbench} with 14 perturbation axes spanning object appearance, scene context, camera pose, and their aggregate. Following HAMSTER~\cite{li2025hamster}, we select 11 front-camera-visible tasks and report success rates over 25 episodes per perturbation.

For real-world manipulation (\textit{RQ3}), we deploy a Franka Panda arm with an external RGB-D camera (Fig.~\ref{fig:4}b) on three task families: \emph{button pressing} (3 buttons), \emph{pouring} (2 cups, 3 bowls), and \emph{pick-and-place} (10 objects, 3 bowls). Each is evaluated under four generalization axes: \emph{language} (unseen synonyms), \emph{spatial} (relative verbal references, and unseen object heights), \emph{visual} (novel textures, lighting, and distractors), and \emph{multiple} (all combined). Rather than binary success, we adopt a continuous score in $[0,1]$: pick-and-place and pouring award 25\% at each of four stages (reach, grasp, transport, place/pour), while button pressing awards 50\% each for positioning and actuation. All results are averaged over a total of 25 evaluation rollouts.

\paragraph{Baselines} For trajectory prediction (\textit{RQ1}), we compare against four proprietary VLMs (Sonnet-4.6, GPT-5.2, Gemini-3.0-Pro) and the open-source RoboBrain-2.5-8B~\cite{tan2026robobrain25}. RoboTracer~\cite{zhou2025robotracer} is excluded as it has not been publicly released. For end-to-end manipulation (\textit{RQ2}, \textit{RQ3}), all methods share the same 3DFA~\cite{gkanatsios20253dfa} low-level policy trained on identical demonstration data, isolating the effect of the guidance signal. We compare: (i)~3DFA without guidance, (ii)~3DFA with 2D trajectories from HAMSTER~\cite{li2025hamster}, and (iii)~3DFA with 3D trajectories from \model{}. For real-world experiments, we additionally include $\pi_{0.5}$~\cite{black2025pi05}, a state-of-the-art monolithic VLA, fine-tuned on the same demonstration dataset as a baseline.

\paragraph{Implementation Details} For the 3D trajectory planner, the VLM backbone is Qwen3-VL-8B-Instruct~\cite{bai2025qwen3vl}, with the depth encoder and decoder initialized from LingBot-Depth~\cite{tan2026lingbotdepth}. Pixel coordinates are normalized to $[0, 1000]$ and depth is predicted in meters. Stage~1 trains the depth projector and decoder only; Stage~2 keeps both encoders frozen and applies LoRA~\cite{hu2022lora} (rank 64) to the LLM while fully training the projectors and decoder. Both stages run for 1 epoch (lr $= 1{\times}10^{-4}$, warmup 0.03, batch 256, $8{\times}$ H100). For the low-level policy, we use 3DFA~\cite{gkanatsios20253dfa} with RGB-D, language, and proprioceptive inputs to predict action chunks of length 20. In simulation, the policy trains on 100 demonstrations per task for 500k steps (batch 64). In the real world, we collect 300/144/108 episodes for pick-and-place/pouring/button pressing and train for 300k steps (batch 64, $4{\times}$ H100). All use lr $= 1{\times}10^{-4}$. For $\pi_{0.5}$~\cite{black2025pi05}, we fine-tune for 50k steps (lr $= 1{\times}10^{-5}$, batch 64), limiting training duration to preserve the base model's pre-trained knowledge.

\begin{figure*}[t]
    \centering
    \includegraphics[width=0.9\textwidth]{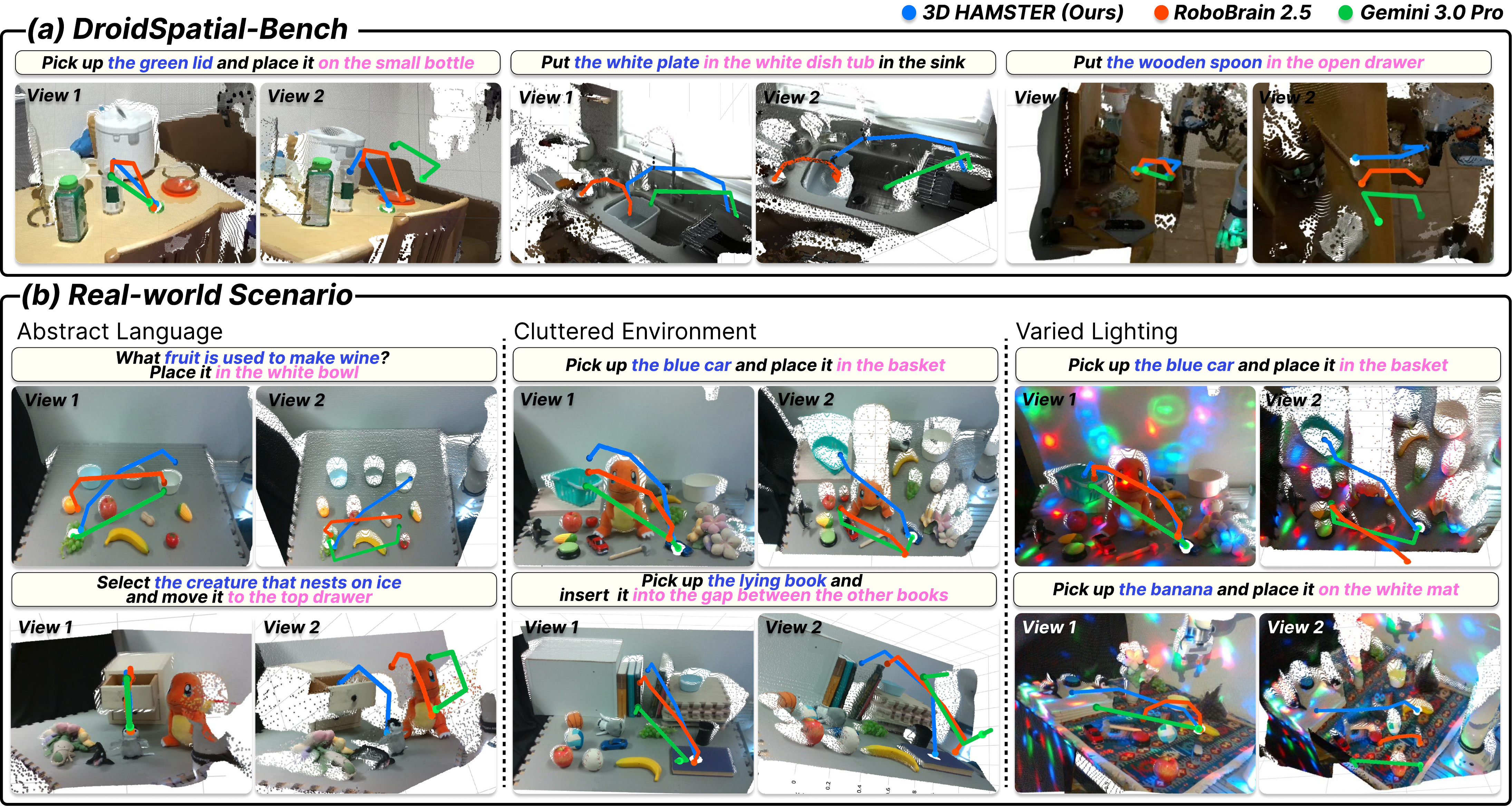}
    \caption{
    % Qualitative results.
    Qualitative comparison of 3D trajectory predictions. Each trajectory is shown from two viewpoints: baseline predictions that appear plausible in View~1 reveal significant depth errors in View~2, whereas \model{} remains metrically consistent across both views.
    }
    \label{fig:3}
    \vspace{-0.2cm}
\end{figure*}

\begin{table}[t]
\centering
\caption{3D Trajectory prediction accuracy (\%) on \benchmark{}. \textbf{Bold} = best per column.}
\label{tab:traj_prediction}
\small
\setlength{\tabcolsep}{2.5pt}
\begin{tabular}{@{}ll ccc c ccc@{}}
\toprule
& & \multicolumn{3}{c}{$\delta = 5$\,cm} & \phantom{a} & \multicolumn{3}{c}{$\delta = 10$\,cm} \\
\cmidrule(lr){3-5} \cmidrule(lr){7-9}
\multirow{-2}{*}{\textbf{Model}} & \multirow{-2}{*}{\textbf{Input}}
& Start & End & Both && Start & End & Both \\
\midrule
\multicolumn{9}{@{}l}{\textit{Proprietary API models}} \\[2pt]
Sonnet-4.6 & RGB
& 1.4 & 8.8 & 0.7 && 6.1 & 16.2 & 2.0 \\
GPT-5.2 & RGB
& 6.8 & 29.7 & 2.7 && 29.7 & 45.3 & 16.2 \\
Gemini-3.0-Pro & RGB
& 29.1 & 44.0 & 16.2 && 43.2 & 56.1 & 29.7 \\
\midrule
\multicolumn{9}{@{}l}{\textit{Open-source models}} \\[2pt]
RoboBrain-2.5-8B & RGB
& 61.5 & 58.1 & 39.2 && 80.4 & 74.3 & 60.1 \\
\midrule
\multicolumn{9}{@{}l}{\textit{\textbf{\model{}} (ours)}} \\[2pt]
Qwen3-VL-8B & RGB
& 0.7 & 9.5 & 0.7 && 0.7 & 14.9 & 0.7 \\
+ 3D Traj. Data & RGB
& 50.0 & 50.0 & 27.7 && 71.6 & 72.3 & 50.0 \\
+ Depth encoder & RGBD
& 62.8 & 62.2 & \textbf{42.6} && \textbf{83.8} & 75.0 & 62.8 \\
\rowcolor[gray]{0.92}
+ $\mathcal{L}_{\text{depth}}$ & RGBD
& \textbf{63.5} & \textbf{66.2} & 41.9 && 80.4 & \textbf{82.4} & \textbf{65.5} \\
\bottomrule
\end{tabular}
\vspace{-0.5cm}
\end{table}

\begin{table*}[t]
\centering
\caption{Per-variation success rates (\%) averaged across tasks on Colosseum~\cite{pumacay2024colosseum}. \textbf{Bold} = best per column.}
\label{tab:colosseum}
\small
\setlength{\tabcolsep}{4pt}
\begin{tabular}{lccccccccccc}
\toprule
Method & None & MO & RO & Light & Table & Distract & BG Tex & RLB Var & Cam Pose & All Var & Avg. \\
\midrule
3DFA & 53.8 & 39.0 & 27.2 & 38.0 & 30.3 & 36.4 & 43.3 & 50.0 & 46.8 & 0.8 & 36.6 \\
3DFA + HAMSTER & 49.5 & 38.1 & 28.8 & 38.8 & \textbf{42.3} & 40.8 & 43.3 & 50.5 & 48.4 & \textbf{7.2} & 38.8 \\
\rowcolor[gray]{0.92} \textbf{3DFA + \model{}} & \textbf{62.9} & \textbf{49.4} & \textbf{36.3} & \textbf{54.4} & 39.6 & \textbf{44.8} & \textbf{52.0} & \textbf{52.5} & \textbf{49.2} & \textbf{7.2} & \textbf{44.8} \\
\bottomrule
\end{tabular}
\vspace{-0.45cm}
\end{table*}

% ============================================================
% RQ1
% ============================================================
\subsection{RQ1: 3D Trajectory Prediction Accuracy}
\label{sec:exp:rq1}

\paragraph{Depth-augmented VLM outperforms proprietary and open-source baselines}
Table~\ref{tab:traj_prediction} reports trajectory prediction accuracy on \benchmark{}. Proprietary VLMs perform poorly at 3D metric trajectory prediction despite their strong general vision-language capabilities: even the best, Gemini-3.0-Pro, achieves only 16.2\% on the strict 5\,cm \emph{Both} metric, confirming that off-the-shelf VLMs lack metric depth reasoning. RoboBrain-2.5-8B provides a much stronger comparison point, reaching 39.2\% (5\,cm \emph{Both}) and 60.1\% (10\,cm \emph{Both}) with RGB input alone. However, \model{} surpasses it across all metrics, with the largest improvements on end-position accuracy (66.2\% vs.\ 58.1\% at $\delta{=}5$\,cm; 82.4\% vs.\ 74.3\% at $\delta{=}10$\,cm), where accurate depth prediction is most critical for downstream placement success.

\begin{figure*}[t]
    \centering
    \includegraphics[width=\linewidth]{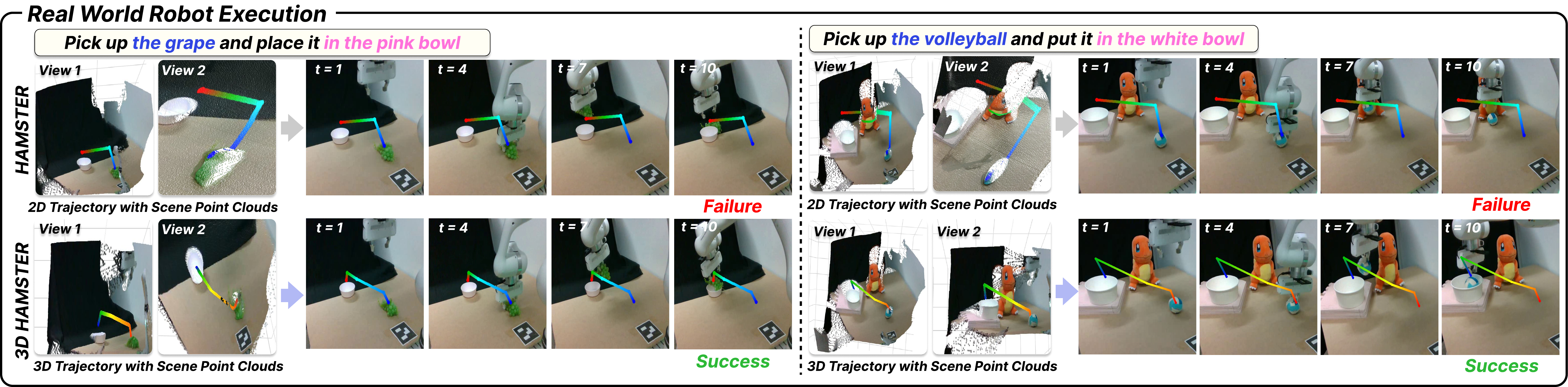}
    \caption{Real-world execution rollouts. HAMSTER's 2D trajectories cling to the scene surface (top row, graffiti effect), causing failure. \model{}'s 3D trajectories remain metrically reliable (bottom row), enabling success.}
    \label{fig:5}
    \vspace{-0.55cm}
\end{figure*}

\paragraph{Training data, depth encoding, and reconstruction each provides gain}
The bottom block of Table~\ref{tab:traj_prediction} isolates each component by incrementally building upon Qwen3-VL-8B.
(i) \emph{3D trajectory dataset supervision} based on our multi-task mixture data provides a substantial boost (5\,cm \emph{Both}: 0.7\%$\to$27.7\%), confirming that targeted training is required to channel the base VLM's general 3D knowledge into effective trajectory prediction. \emph{(ii) Depth encoder} yields further substantial gains (5\,cm \emph{Both}: 27.7\%$\to$42.6\%, 10\,cm \emph{Both}: 50.0\%$\to$62.8\%), indicating that RGB alone struggles in reliably recovering metric depth. \emph{(iii) Depth reconstruction loss} improves end-point accuracy (10\,cm \emph{End}: 75.0\%$\to$82.4\%), where depth drift across waypoints is greatest. This suggests that sparse trajectory supervision alone cannot maintain depth fidelity over the full sequence; the dense reconstruction loss counteracts this drift by anchoring depth representations to a scene-level geometric prior.

\paragraph{Qualitative results for in- and out-of-distribution scenes}
Fig.~\ref{fig:3} visualizes predicted trajectories from \model{}, RoboBrain~2.5, and Gemini~3.0~Pro. On \benchmark{} (Fig.~\ref{fig:3}a), baseline predictions appear reasonable from View~1 but reveal substantial depth errors when viewed from a second angle, whereas \model{} remains metrically consistent across both viewpoints. This gap widens in Fig.~\ref{fig:3}b, where all models are evaluated zero-shot on real-world scenes unseen during training. Baseline trajectories lose spatial coherence entirely, while \model{} continues to produce geometrically plausible paths, demonstrating that the depth-augmented planner generalizes its metric 3D reasoning beyond the training distribution.

% ============================================================
% RQ2
% ============================================================
\subsection{RQ2: Robustness Under Systematic Distribution Shifts}
\label{sec:exp:rq2}

\paragraph{2D guidance harms in-distribution performance; 3D guidance improves it}
Table~\ref{tab:colosseum} shows that \model{} achieves the highest average success rate (44.8\%), improving over 2D guidance (HAMSTER) by 6.0\% and over the unguided baseline by 8.2\%. Notably, 2D guidance actually \emph{degrades} unperturbed performance relative to no guidance (49.5\% vs.\ 53.8\%), indicating that projecting plans into 2D pixel coordinates introduces geometric ambiguity that harms the 3D low-level controller even on in-distribution data. In contrast, 3D guidance improves the unperturbed condition to 62.9\%, confirming that metrically reliable trajectories align naturally with the policy's 3D action space.

\paragraph{3D guidance provides both appearance invariance and geometric robustness}
3D guidance yields the largest gains over 2D on lighting (+15.6\%), manipulated object (MO, +11.3\%), and background texture (BG Tex., +8.7\%) perturbations. These axes directly corrupt the color and texture features that 2D planners depend on for implicit depth inference, whereas 3D trajectories specify the path in metric coordinates independently of visual appearance. Notably, MO and RO perturbations in Colosseum alter not only object color and texture but also object size, meaning the task geometry itself changes. That 3D guidance still outperforms 2D under these combined appearance and geometry shifts suggests its benefit extends beyond appearance invariance, providing metrically stable waypoints even under geometric variation. However, under the All Var.\ condition, where all perturbations compound simultaneously, both guided methods achieve only 7.2\%. While this still substantially exceeds the unguided baseline (0.8\%), the 2D and 3D planners converge, indicating that when severe visual corruption overwhelms the VLM's ability to ground the target object, guidance dimensionality becomes irrelevant.

% ============================================================
% RQ3
% ============================================================
\subsection{RQ3: Real-World Transfer Across Generalization Axes}
\label{sec:exp:rq3}

\paragraph{3D guidance consistently improves real-world success, especially under out-of-distribution shifts}
Table~\ref{tab:realworld} summarizes real-world results. \model{} achieves the highest average success across all three task families (80\% button pressing, 68\% pouring, 62\% pick-and-place), outperforming both 2D guidance (HAMSTER: 60\%, 45\%, 46\%) and the monolithic $\pi_{0.5}$ (74\%, 41\%, 40\%). While $\pi_{0.5}$ matches or exceeds \model{} in-distribution (e.g., 100\% on button pressing and pick-and-place), it degrades sharply under distribution shifts, dropping to 15\% on pouring and 15\% on pick-and-place under spatial variation. This confirms that monolithic VLAs overfit to training conditions despite strong in-distribution performance. The advantage of 3D over 2D guidance is most pronounced under visual shifts (button pressing: 100\% vs.\ 80\%; pouring: 65\% vs.\ 35\%) and spatial shifts (button pressing: 50\% vs.\ 20\%; pouring: 65\% vs.\ 40\%), where implicit depth inference from altered appearance breaks down. The largest overall gain appears in pouring (68\% vs.\ 45\%), which demands precise height control when tilting the cup, a geometric requirement that 3D waypoints directly encode. Under compounding shifts (multiple), 3D guidance degrades gracefully across all tasks, consistently outperforming or matching all baselines.

\begin{table}[t]
\centering
\caption{Real-world success rates (\%). \textbf{Bold} = best per cell.}
\label{tab:realworld}
\small
\setlength{\tabcolsep}{3pt}
\begin{tabular}{@{}l ccccc c@{}}
\toprule
& In-D & Lang & Spat & Vis & Mult & Avg. \\
\midrule
\multicolumn{7}{@{}l}{\textit{Button Pressing}} \\[2pt]
$\pi_{0.5}$             & \textbf{100} & 80 & 40 & 90 & 60 & 74 \\
3DFA                    & 90 & 30 & 0 & 30 & 40 & 38 \\
3DFA + HAMSTER          & 80 & 50 & 20 & 80 & \textbf{70} & 60 \\
\rowcolor[gray]{0.92} \textbf{3DFA + \model{}}  & \textbf{100} & \textbf{90} & \textbf{50} & \textbf{100} & 60 & \textbf{80} \\
\midrule
\multicolumn{7}{@{}l}{\textit{Pouring}} \\[2pt]
$\pi_{0.5}$             & 60 & 15 & 35 & \textbf{65} & 30 & 41 \\
3DFA                    & 80 & 45 & 50 & 50 & 25 & 50 \\
3DFA + HAMSTER          & 75 & 45 & 40 & 35 & 30 & 45 \\
\rowcolor[gray]{0.92} \textbf{3DFA + \model{}}  & \textbf{95} & \textbf{75} & \textbf{65} & \textbf{65} & \textbf{40} & \textbf{68} \\
\midrule
\multicolumn{7}{@{}l}{\textit{Pick-and-Place}} \\[2pt]
$\pi_{0.5}$             & \textbf{100} & 35 & 15 & 20 & 30 & 40 \\
3DFA                    & 65 & 45 & 5 & 25 & 10 & 30 \\
3DFA + HAMSTER          & 70 & 75 & 35 & 15 & 35 & 46 \\
\rowcolor[gray]{0.92} \textbf{3DFA + \model{}}  & 90 & \textbf{95} & \textbf{40} & \textbf{35} & \textbf{50} & \textbf{62} \\
\bottomrule
\end{tabular}
\vspace{-0.6cm}
\end{table}

\paragraph{3D guidance executes the planner's intent faithfully; 2D guidance distorts it}
Fig.~\ref{fig:5} compares real-world execution rollouts of HAMSTER and \model{} on two pick-and-place tasks. For HAMSTER, the 2D trajectory appears plausible from View~1 but View~2 reveals the \textit{graffiti effect}: since the 2D planner provides no depth, each waypoint is assigned the depth of the underlying scene surface, pinning the trajectory to the point cloud rather than tracing a free path through 3D space. The low-level policy cannot distinguish this surface-bound guidance from the scene geometry itself, resulting in task failure in both examples. In contrast, \model{}'s 3D trajectories remain metrically consistent across both viewpoints, clearly separated from the scene surface and faithfully encoding the intended pick-and-place motion. The low-level policy tracks these waypoints accurately, achieving successful execution in both tasks. This comparison directly shows the core advantage of operating in a shared 3D metric space: the controller receives guidance that exists in its own coordinate frame, eliminating the geometric distortion inherent in 2D-to-3D lifting.

% ============================================================
% Conclusion
% ============================================================
\section{CONCLUSION}
\label{sec:conclusion}

We presented \model{}, a hierarchical framework that aligns high-level planning and low-level control in metric 3D space. A pretrained VLM augmented with a depth encoder and a dense depth reconstruction loss generates metrically reliable $(u,v,d)$ trajectories, which are directly integrated into a pointcloud-based low-level policy. Our experiments show that (1) each component of the depth-augmented planner provides complementary gains in 3D trajectory prediction accuracy, outperforming Gemini-3.0-Pro and RoboBrain-2.5 on \benchmark{}; (2) 3D guidance consistently outperforms 2D alternatives in simulation, with the largest improvements under perturbations that corrupt visual appearance or alter object geometry; and (3) these benefits transfer to real-world manipulation, where 3D guidance improves performance across language, spatial, and visual generalization axes, with the most substantial gains in tasks demanding precise depth control such as pouring.
Several limitations suggest directions for future work. First, our framework requires explicit depth maps from an RGB-D sensor; integrating monocular depth estimation would remove this hardware dependency. Second, the planner operates from a single viewpoint, making it vulnerable to heavy occlusion; multi-view fusion would improve robustness in cluttered scenes. Finally, our evaluation is limited to single-arm tabletop tasks; extending to mobile manipulation and bimanual coordination would test the generality of 3D trajectory guidance in more complex settings.

\bibliographystyle{IEEEtran}
\bibliography{main}

\end{document}